\documentclass[journal]{IEEEtran}

\usepackage{cite}
\usepackage{graphicx}
\ifCLASSOPTIONcompsoc
\usepackage[caption=false,font=normalsize,labelfont=sf,textfont=sf]{subfig}
\else
\usepackage[caption=false,font=footnotesize]{subfig}
\fi

\usepackage{url}
\usepackage{amsmath}
\usepackage{amssymb}
\usepackage{amsbsy}
\usepackage{array}

\usepackage{epstopdf}

\newcommand{\mean}[1]{\ensuremath{\langle #1 \;\rangle}}

\begin{document}
%
\title{Evaluating Registration Without Ground Truth}
%
%
\author{Carole J. Twining, Vladimir S. Petrovi\'{c}, Timothy F. Cootes, Roy S. Schestowitz, William~R.~Crum, and Christopher J. Taylor
\thanks{This research was supported by the \mbox{MIAS} \mbox{IRC} project, \mbox{EPSRC}
grant No. \mbox{GR/N14248/01}, UK Medical Research Council Grant
No. \mbox{D2025/31}, and also by the \mbox{IBIM} project,
\mbox{EPSRC} grant No. \mbox{GR/S82503/01}. \textit{(Corresponding author: Carole J. Twining.)}}
\thanks{W. R. Crum is in the Department of Surgery \& Cancer, Faculty of Medicine, Imperial College London, South Kensington Campus, London SW7 2AZ, United Kingdom. All other authors
are with the Division of Informatics, Imaging and Data Sciences, University of Manchester, Manchester M13 9PT, United Kingdom. Email: carole.twining@manchester.ac.uk}
}


\markboth{[DRAFT Placeholder] Transactions on Medical
Imaging,~Vol.~, No.~,~~00~~}{Someone \MakeLowercase{\textit{et
al.}}: IEEE TMI Paper}

%

\IEEEpubid{0000--0000/00\$00.00˜\copyright˜2020 IEEE}



\maketitle

\begin{abstract}
We present a generic method for assessing the quality of
non-rigid registration~(NRR) algorithms, that {\em does not}
depend on the existence of any ground truth, but depends solely on
the data itself. The data is a set of images. The
output of any NRR of such a set of images is a
dense correspondence across the whole set. Given such a dense
correspondence, it is possible to build various generative
statistical models of appearance variation across the set.
We show that evaluating the quality of the registration can be mapped to the problem of evaluating the quality of the resultant statistical model. The quality of the model entails a comparison between the model and the image data that was used to construct it. It should be noted that
this approach does not depend on the specifics of the registration
algorithm used (i.e., whether a groupwise or pairwise algorithm
was used to register the set of images), or on the specifics of
the modelling approach used.

We derive an index of image model specificity that can be used to assess image model quality, and hence the quality of registration. This approach is validated by comparing our assessment of registration quality
with that derived from ground truth anatomical labeling. We
demonstrate that our approach is capable of assessing NRR reliably
without ground truth. Finally, to demonstrate the practicality of
our method, different NRR algorithms -- both pairwise and
groupwise -- are compared in terms of their performance on 3D MR
brain data.
\end{abstract}

%
\IEEEpeerreviewmaketitle


\section{Introduction}
\label{sec:Introduction}

\IEEEPARstart{N}{on-rigid} registration (NRR) of images has been used extensively in recent years, as a basis
for medical image analysis across a wide-range of applications in the medical field~\cite{Oliveira_Review}.
The problem is highly under-constrained and a plethora of different
algorithms have been proposed. The aim of NRR is to find, automatically, a
meaningful, dense correspondence across a pair (hence {\em
pairwise} registration), or group (hence {\em groupwise}) of
images. A typical algorithm consists of three main components: transformation, similarity, and optimization. The transformation is a representation of the deformation fields that encode the spatial relationships between images. The similarity measure is an objective function that quantifies the degree of (mis-)registration, which is optimized to achieve the final result. Different choices of these components result in different algorithms, and when these algorithms are applied to the same data the results typically differ~\cite{Zitova_2003}. There is thus a need for quantitative methods to compare the quality of registration produced by different algorithms (allowing the best method to be selected and any parameters tuned). Ideally it would be possible to establish the best choice for a given data set, since results may be data-dependent.

\IEEEpubidadjcol

Numerous methods have been proposed for assessing the results of
NRR~\cite{Hellier,Validation-NRR,Schnabel,Klein2009,Klein2010}. One obvious approach
is to compare the results of the registration with anatomical ground
truth, where it is available. This, however, requires expert annotation which is labour-intensive to generate (particularly in 3D), subjective and impractical to obtain for every data-set of interest. An alternative approach is to synthesize artificial test data for which the ground truth registration is known by construction.  In its simplest form this involves generating multiple artificial distortions of an example image.  The weakness of this approach is that it is difficult to guarantee that the synthetic image set is properly representative of real data.  These weaknesses motivate the search for a method of evaluation that does not depend on the
existence of ground truth data, or on our ability to synthesize realistic image data variation, but which uses the data to be registered, itself, as the basis for evaluation.

The method we will describe achieves this by drawing on ideas from statistical modelling of object appearance (shape and texture), as used extensively for interpretation by synthesis (e.g., Active Appearance Models (AAMs)~\cite{Cootes_2001}).  The well-known connection is that building a statistical appearance model from a set of images involves establishing a dense correspondence between them, whilst the desired output of a successful NRR algorithm is such a dense correspondence. Thus the output of any NRR algorithm can be used to build a model, different NRR algorithms will give different models, and the `best' registration will give the `best' model. The registration and modelling aspects can hence be combined or interleaved, to produce either a groupwise registration algorithm such as the Minimum Description Length~(MDL) registration algorithm~\cite{IPMI_2005_ISBE, Davies}, or to construct generative shape and appearance models from images without annotation, as demonstrated by Ashburner at al.~\cite{Ashburner2019}. As noted by Ashburner et al., the use of such generative image models can also be related to the generative approaches to machine learning, such as the family of generative adversarial networks~\cite{goodfellow2014generative}.

We are guided by both of these observations in the current paper, in that we map the problem of evaluating the quality of a general NRR to evaluating the quality of the model generated using it, and also that the \textit{generative} aspect of the statistical image model is a key aspect of the method of evaluation.

The structure of the paper is as follows. Section~\ref{sec:Background} gives a brief background to both the assessment of NRR, and of the construction of appearance models, explaining in more detail the link between the two. Section ~\ref{sec:Evaluation} details our
method for obtaining `ground-truth-free' quantitative measures of registration quality, and we present results of extensive validation experiments (Section~\ref{sec:Validation}), comparing, under controlled conditions, the behavior of our measure with an established measure based on anatomical ground truth. In Section~\ref{sec:Experiments} the method is applied to evaluate the results obtained using several different NRR algorithms to register a set of 3D MR brain images, again demonstrating consistency with evaluation based on ground truth. We conclude with a discussion in Section~\ref{sec:Conclusions}.

\section{Background}\label{sec:Background}

\subsection{Non-Rigid Registration}

The aim of NRR is to find an anatomically
meaningful, dense (i.e., pixel-to-pixel or voxel-to-voxel)
correspondence across a set of images. This correspondence is
typically encoded as a spatial deformation field between each
image and a reference image, so that when one is deformed onto the other,
corresponding structures are brought into alignment. This is
generally a challenging problem,
due to the extent and complexity of cross-individual
anatomical variation. In addition, exact structural
correspondence may not exist between images, or the class of
spatial deformation fields employed may not be able to represent the
correct correspondence exactly.

\subsection{Assessment of Non-Rigid
Registration}\label{Odef} We describe several commonly-used
approaches to the problem of assessing the results of
registration.

\subsubsection{\bf Recovery of Deformation Fields}
One obvious way to test the performance of a registration
algorithm is to apply it to some {\em artificial} data where the
true correspondence is known. Such test data is typically
constructed by applying a set of known deformations
to a real image. This artificially-deformed data is
then registered, and evaluation is based on comparison
between the deformation fields recovered by the NRR and
those that were originally
applied~\cite{Validation-NRR,Schnabel}. This approach has two
limitations: (i) since the images to be registered are
derived from the same source image, they are structurally similar
and have similar appearance; (ii) it is difficult to ensure that
the deformations used to generate the test images are typical of
the anatomical variability found in real data.  Together, these
factors lead to an NRR problem that is not necessarily representative of
real data.

\subsubsection{\bf Overlap-based Assessment} The
overlap-based approach involves measuring the overlap of
anatomical annotations before and after registration. Examples of
this approach involve measurement of the misregistration of
anatomical regions of significance~\cite{Hellier}, and the overlap
between anatomically equivalent regions obtained using
segmentation. This process is either manual or
semi-automatic~\cite{Hellier,Validation-NRR}. Although these
methods cover a general range of applications, they are
labour-intensive, and the manual component often suffers
from excessive subjectivity.

In some applications, the exact dense details of the deformation
fields are of intrinsic interest, and the ability of a
registration algorithm to accurately recover them should be
tested. In other applications, it is of more importance that
structural and functional correspondence is achieved, which
requires that voxel-wise correspondence in terms of structural
labels should be tested directly. This paper utilizes, as a \textit{reference} for comparison purposes, one such
method, which assesses registration using the spatial overlap. In
imaged objects with multiple labels, a simple overlap measure can
be used to assess overlap on a structure-by-structure basis. But
combining the overlap assessment from multiple labels requires a
rather more sophisticated approach, as follows.

\subsubsection{\bf The Generalized Overlap}
For single structures, overlap is defined using the standard Jaccard/Tanimoto~\cite{Jaccard,Tanimoto,Tani2} formulation, for corresponding regions in the registered images. The correspondence
is defined by labels of distinct image regions (in this case brain
tissue classes), produced by manual mark-up of the original images
(ground truth labels). A correctly registered image set will exhibit high relative overlap between corresponding brain structures in different images and, in the opposite case -- low
overlap with non-corresponding structures. Although, for simplicity, it can be assumed that the ground truth labels are strictly binary, the same is not true after registration and re-sampling. Our main focus is assessment that requires no ground truth, but the approach above provides a good reference to compare
against for validity with respect to ground truth annotation. Overlap for multiple structures are computed using the generalized overlap measure of Crum et al.~\cite{Crum}, which outputs a single figure of merit for the overlap of all labels over all subjects. The relative weighting of different labels allows for further tuning of the overlap measure. Our measure is equation (4) from Crum et al.~\cite{Crum}, but without the pairwise weights.

\subsection{Statistical Models of Appearance}
We focus on the classic generative appearance models
of Cootes et al.~\cite{Cootes_ECCV_1998,Edwards}, though other
approaches could equally well have been explored.  Such models continue to be used extensively across a wide range of medical image analysis applications, either individually~(e.g. \cite{de2019quantifying,reyneke2018review}), or combined with deep learning refinements~\cite{cheng2016active}.

These appearance models capture complex variation in both shape and
intensity (texture), are constructed from training sets of example
images, and require a dense correspondence to be established
across the set.  In the current setting, these correspondences will
be provided by the results of NRR ~\cite{Frangi}. Model-building
starts with a training set of affinely-aligned images. The results of NRR on this set
can be expressed as a vector $\vec{x}$ for each image, with elements
that are the positions in the image corresponding to a selected set
of points in the reference (eg. every pixel/voxel, or a set of landmark
points sufficient to specify the deformation field).  Similarly, a
shape-free texture vector $\vec{g}$ can be formed for each image by
sampling the intensity at a set of points corresponding to
regularly spaced points in the reference.

In the simplest case, we model the variation of shape and texture
in terms of multivariate Gaussian distributions, using Principal
Component Analysis (PCA)
, leading to linear statistical models of the form:
\begin{equation}
\vec{x}= \mean{\vec{x}}+\mathbf{P}_{s}\vec{b}_{s}, \:\:\:\:
\vec{g}= \mean{\vec{g}}+\mathbf{P}_{g}\vec{b}_{g}, \label{eq:AppModel}
\end{equation}
where $\vec{b}_{s,g}$ is the vector of shape/texture parameters, and $\mean{\cdot}$ is the corresponding mean. The matrices $\mathbf{P}_{s}$ and $\mathbf{P}_{g}$ encode the
principal modes of shape and texture variation respectively.

If the variations of shape and texture are correlated, we can also obtain a
combined model of the more general form:
\begin{equation}
\vec{x} = \mean{\vec{x}}+\mathbf{Q}_{s}\vec{c}, \:\:\:\:
\vec{g} = \mean{\vec{g}}+\mathbf{Q}_{g}\vec{c}.\label{eq:ComAppModel}
\end{equation}
The model parameters $\vec{c}$ control both shape and
texture, and $\mathbf{Q}_{s}$, $\mathbf{Q}_{g}$ are matrices
describing the general modes of variation derived from the
training set. New images from the distribution from which the
training set were drawn can be generated by varying $\vec{c}$.
For further implementation details for such
appearance models, see~\cite{Cootes_ECCV_1998,Edwards}.

\section{Evaluating NRR Quality}
\label{sec:Evaluation}

We here discuss methods for assessing
the quality of the model built from the results of NRR, and hence the quality of
the registration, with the aim of finding one that
is both robust and sensitive to small changes in registration quality.

\subsection{Specificity and Generalization}

Given a training set of image examples, the aim of modelling is to
estimate the probability density function (pdf) of the process
from which the training set was sampled. Ideally, evaluation
of the model would involve comparing this estimate with the true pdf
-- except, of course, that is not possible, since we only have access
to the training set, not the underlying distribution. We cannot use measures that may be used to drive the registration, since these will be optimal by construction and hence will not provide an unbiased measure of model quality. Given these constraints, we follow the approach of Davies et al.
~\cite{IPMI_2005_ISBE,Davies}, who introduced the concepts
of {\em specificity} (and the related concept of {\em generalization}) when comparing the
performance of different model-building algorithms.
It has been shown that these measures can be
considered as graph-based estimators of the cross-entropy of the
training set distribution and the model distribution~\cite{US:paper,US:Book}. This provides
a firm theoretical justification for their use, as well-defined
measures of overlap between two distributions.

For models, {\em generalization ability} refers to the ability to interpolate and extrapolate from the training set, to generate novel examples similar to the training data. Whilst the converse concept of {\em specificity} refers to the ability to only generate examples which can be considered as valid examples of the class of objects being imaged.

In order to construct quantitative measures based on these concepts, we consider our training set of examples
as a set of points in some data space $\mathbb{R}^{m}$. The exact
nature and dimensionality of this data space depends on our choice
of {\em representation} of the training data. Let
\mbox{$\mathcal{X} \doteq \{\vec{x}_{i} \in
\mathbb{R}^{m}:i=1,\ldots N\}$} denote the  $N$ examples of the
training set when considered as points in this data space. Our
statistical model of the training set is then a pdf defined in the
same space, \mbox{$p(\vec{z}), \:\: \vec{z} \in \mathbb{R}^{m}$.}

The quantitative measure of the {\em specificity} $S$ of the model
with respect to the training set $\mathcal{X} = \{\vec{x}_{i}\}$
is then given by:
\begin{equation}
\hat{S}_{\lambda}(\mathcal{X};p) \doteq \int p(\vec{z})
\underset{i}{\min}\left(|\vec{z}-\vec{x}_{i}|\right)^{\lambda} \:
d\vec{z},\label{eq:Sdef}
\end{equation}
where $|\cdot|$ is a distance on our data space, and
$\lambda$ is a positive number. That is, for each point $z$ in the data space, we compute the sum of powers of NN-distances, weighted by the pdf $p(z)$. Thus large values of
$S$ correspond to model distributions that extend beyond the training
set, and have poor specificity, whereas small values of $S$
indicate models with better specificity.

This integral form of the specificity can be approximated using a
Monte-Carlo method, as follows. We
generate a large random set of points in data space
\mbox{$\mathcal{Y} = \{\vec{y}_{\mu}:\, \mu=1,\ldots M\}$}, sampled from the model pdf $p(\vec{z})$. The estimate of (\ref{eq:Sdef}) is then:
\begin{equation}\label{Seq}
\hat{S}_{\lambda}(\mathcal{X};p)\approx
S_{\lambda}(\mathcal{X,Y}) \doteq
\frac{1}{M}\sum\limits_{\mu=1}^{M}\underset{i}{\min}\left(|\vec{x}_{i}-\vec{y}_{\mu}|\right)^{\lambda},
\end{equation}
and we call $S_{\lambda}$ the measured
specificity. The standard error is generated from the standard deviation of
the measurements:
\begin{equation}
\sigma_{S}=\frac{1}{\sqrt{M-1}} \:\: \underset{\mu=1\ldots M}{\mathrm{STD}}
\left\{\underset{i}{\min}\{|\vec{x}_{i}-\vec{y}_{\mu}|^\lambda\}\right\}.
\end{equation}
Generalization is defined in an analogous
manner, by just swapping the roles of training points
$\mathcal{X}$ and the sample points $\mathcal{Y}$. The standard error on the specificity depends on the
number $M$ of elements in the sample set, which (computational time permitting), can be made as large as required, hence the standard error on the specificity can be easily reduced.
In contrast, the error on the generalization depends on the number
of elements $N$ in the training set, which cannot be varied. Thus
specificity is likely to be a more reliable measure than generalization for small training sets, and is the measure we will use in the rest of this paper.

\subsection{Specificity for Image Models}\label{sec:ImageModels}

In earlier preliminary work~\cite{Roy},
the registration was evaluated by computing the specificity of the
full appearance model built from the registered images. This gave
promising results on 2D image sets, but there are various theoretical
and practical problems that make this unsuitable for large-scale evaluation of registration on 3D image sets. Specificity has mainly been used to evaluate various types of \textit{shape} model (e.g., Tu et al.~\cite{tu2017skeletal} evaluated 3D skeletal S-REPS compared with various types of point-based models (PDMs)), but the application to image-based models is rather different, and requires the consideration of several issues that are not present in the case of shape models, as we will now show.

Let us first consider the question of computing the specificity of a full appearance model. Let $\{I_{i}:  i=1,\ldots N\}$ represent
the $N$ images in the training set, each with $n$ pixels/voxels, which we will suppose have
been rigidly or affinely aligned, hence eliminating the degrees of
freedom associated with variation in pose. Each training image is then represented by a vector
$\vec{I}_{i} \in \mathbb{R}^{n}$, with components
\mbox{$\{I_{i}(\vec{p}_{a}):a=1,\ldots n\}$}, where
$\{\vec{p}_{a}\}$ are the positions of all pixels/voxels in the
images.
\subsubsection{\bf The Appearance Model Sub-Space}
Each training image is represented in the model~((\ref{eq:AppModel}) or (\ref{eq:ComAppModel})) by applying texture then shape variations to the model reference image. The texture model is constructed by warping each training image into the spatial frame of the reference. Because of the image pixellation, some information is {\em always} lost in this process, when re-sampling. This means that even if we retain {\em all} of the
modes of shape-free texture variation, we still do not get perfect reconstruction of our training images
when we warp from the reference frame back to the frame of a
particular training image.

In terms of our data space $\mathbb{R}^{n}$ of affinely aligned
images, this means that the appearance model does not span the
Euclidean sub-space $\mathbb{R}^{N-1}$ spanned by the training data. Instead
the model lies on some sub-space $\mathcal{M}$ of $\mathbb{R}^{n}$, where the training images do
not all lie on this sub-manifold. The differences between the
training images and the closest point on the model sub-manifold
$\mathcal{M}$ is a measure of the representation error associated
with each training image. Let us consider evaluating the specificity of the full appearance
model:
\begin{eqnarray}
S_{\lambda}(\mathcal{X,Y}) &\doteq&
\frac{1}{M} \sum\limits_{\mu=1}^{M}\underset{i}{\min} \left(|\vec{I}_{i}-\vec{I}_{\mu}|\right)^{\lambda}, \label{eq:SpecSet} \\
&\approx& \sum\limits_{i=1}^{N} \int\limits_{\Omega(\vec{I}_{i})\cap\mathcal{M}} p(\vec{z})\left(|\vec{I}_{i}-\vec{z}|\right)^{\lambda} d \vec{z},
\end{eqnarray}
where \mbox{$\Omega(\vec{I}_{i}) \doteq
\{\vec{x}\in\mathbb{R}^{n}: |\vec{x}-\vec{I}_{i}| \le
|\vec{x}-\vec{I}_{j}| \: \forall \: j =1,\ldots N\}$} is the
Voronoi cell about the training image $\vec{I}_{i}$, and
$\Omega(\vec{I}_{i})\cap\mathcal{M}$ is the intersection of this
Voronoi cell with the appearance model sub-manifold $\mathcal{M}$.
We can now see that there are several possible problems with
measuring this specificity. The first is that the distance
measurements actually include two different contributions, the
representation error for each training image, then the
contribution to the distance that comes from the distribution of
the training set versus the model distribution. Note that the
separation that is representation error is locally orthogonal to
separations that represent modelled variation. Distances in these
two different sets of directions are not necessarily equivalent.
Hence we see that there are possible problems
involved by subsuming these two different types of variation
within the same simple measure of distance in $\mathbb{R}^{n}$.

The second possible problem is evident if we consider the integral
form of the specificity given above. Not every Voronoi cell
$\Omega(\vec{i})$ necessarily intersects the model sub-space
$\mathcal{M}$, hence not every training image necessarily
contributes to the measured specificity. This problem is
illustrated by the frequency data shown in the upper half of
Table~\ref{tab:AAM}, which considers appearance models built from
3D MR brain data using various non-rigid registrations (as will be
further explained in a later section).
\begin{table*}
\caption{3D brain MR data. The populations of the Voronoi cells of
the training data in the space of images $\mathbb{R}^{n}$, for
$1000$ random examples generated from AAMs built using four
different methods of image registration. The upper half of the
Table shows the results using the Voronoi cells of the raw
training data, whilst the lower half shows the results after
projection of the training data onto the appropriate model
sub-manifold $\mathcal{M}$.} \label{tab:AAM} \centering
\setlength{\tabcolsep}{2pt}.
\begin{tabular*}{\textwidth}{@{\extracolsep{\fill}}|r|*{36}{c}|}
\hline
\bf{Image}&1&2&3&4&5&6&7&8&9&10&11&12&13&14&15&16&17&18&19&20&21&22&23&24&25&26&27&28&29&30&31&32&33&34&35&36\\
\hline
\bf{Fluid}&62&8&\bf{0}&15&\bf{0}&90&8&9&\bf{0}&3&2&23&2&2&138&\bf{0}&67&\bf{0}&1&\bf{0}&3&16&1&53&8&2&29&3&274&17&\bf{0}&16&5&\bf{0}&1&142\\
\bf{Pair1}&117&1&\bf{0}&4&\bf{0}&13&4&2&\bf{0}&2&2&13&\bf{0}&\bf{0}&27&\bf{0}&93&\bf{0}&\bf{0}&\bf{0}&2&1&\bf{0}&2&\bf{0}&\bf{0}&2&1&502&7&\bf{0}&10&3&\bf{0}&\bf{0}&192\\
\bf{Pair2}&106&6&\bf{0}&1&\bf{0}&6&2&3&\bf{0}&3&3&14&\bf{0}&\bf{0}&28&\bf{0}&110&\bf{0}&\bf{0}&\bf{0}&\bf{0}&1&\bf{0}&1&\bf{0}&\bf{0}&\bf{0}&\bf{0}&505&21&\bf{0}&21&3&\bf{0}&\bf{0}&166\\
\bf{Group}&109&4&\bf{0}&4&\bf{0}&6&5&\bf{0}&\bf{0}&1&\bf{0}&3&\bf{0}&\bf{0}&34&\bf{0}&150&\bf{0}&\bf{0}&\bf{0}&1&1&\bf{0}&1&\bf{0}&\bf{0}&2&\bf{0}&496&2&\bf{0}&9&7&\bf{0}&\bf{0}&165\\
\hline
\bf{Image}&1&2&3&4&5&6&7&8&9&10&11&12&13&14&15&16&17&18&19&20&21&22&23&24&25&26&27&28&29&30&31&32&33&34&35&36\\
\hline
\bf{Fluid}&40&25&47&36&6&22&23&13&2&32&38&14&34&35&100&15&66&15&8&34&21&4&1&31&28&10&15&18&72&28&4&71&1&31&16&44\\
\bf{Pair1}&29&30&25&23&14&1&56&5&2&19&26&55&36&87&117&30&8&\bf{0}&11&42&60&6&12&118&26&4&4&11&22&11&2&15&11&37&5&40\\
\bf{Pair2}&48&11&30&13&11&33&50&22&1&37&30&1&63&31&73&55&82&1&30&40&69&3&5&73&10&14&9&8&22&11&9&29&18&16&3&39\\
\bf{Group}&31&44&46&23&9&8&61&7&2&14&62&34&24&20&125&31&32&3&4&46&50&16&1&50&42&22&20&15&33&4&4&53&22&19&1&22\\
\hline
\end{tabular*}

\end{table*}
As we can see from the Table, without projection into the
sub-manifold $\mathcal{M}$ in image space $\mathbb{R}^{n}$ (upper
half of the Table), a large number (approximately $40\%$)  of the
training images are not selected at all (or only very
infrequently), suggesting that their Voronoi cells do not have a
significant overlap with the model sub-manifold. Whereas in some
cases, over $50\%$ of the generated examples all chose the {\em
same} training example as their nearest-neighbor. Yet in each
case, the results of the registration, the resultant appearance models, and the
model representations of each training image were judged to be
reasonable upon visual inspection, hence this problem is not due
to a bad model, but an inherent limitation of the modelling
process.

There is a simple and fairly obvious approach to solving both
these problems, which is to project each training image onto its
model representation, replacing it by the closest point on the model sub-manifold $\mathcal{M}$.
Provided the initial representation errors are judged to be small
enough, this then gives a meaningful measurement of specificity to which
all training images contribute. This is the data shown in the lower half of Table~\ref{tab:AAM}.
Compared to the data in the upper half, we see that the
populations of the Voronoi cells are now much more evenly spread, with
only very occasional zeros or values in single digits.

However, there is a further problem with this proposed solution.
The nature of the spatial warping process from reference frame to
training image frame means that, in general, even if we build a
simple linear model on the space of shape-free texture parameters
and warp parameters, this does not mean that the final appearance
model is linear, nor that $\mathcal{M}$ is a Euclidean sub-space
of $\mathbb{R}^{n}$. If we have projected the training images into
the model sub-space, we should really be using geodesic distances
defined within the sub-manifold to measure the specificity.
However, we do not have access to these distances, only the
distances measured in $\mathbb{R}^{n}$. If the model sub-space is
curved, this means that we will systematically under-estimate the
distances used to measure specificity. Nor do we have direct
access to the shape of the model sub-manifold, except via sample
images generated by the appearance model.

\subsubsection{\bf The Set of Registered Images}
There is, however, a simple alternative to the two approaches
considered above. We are not trying to evaluate the
full appearance model per se, but rather the NRR from which the model was generated. The usual aim of NRR is producing  a set of registered images in
which corresponding structures are aligned across the set, hence what counts is this set of \textit{registered} images. The model built from this set is the shape-free texture model, and it is this
model that we will now evaluate.

Specificity evaluation for this model has several advantages. When evaluating the appearance model, we
needed a definition of distance on the space of affinely-aligned
images $\mathbb{R}^{n}$. In preliminary work~\cite{Roy}, the
shuffle distance was used, which was computationally expensive.
The Euclidean distance between images in $\mathbb{R}^{n}$ is less
computationally expensive. However, we still have to perform the
Voronoi cell and distance computations for vectors of length $n$.
But the shape-free texture model is built on the space of
non-rigidly registered images, rather than the space of affinely-aligned
images. By performing dimensional-reduction (e.g., by using PCA),
we can work instead in the Euclidean sub-space $\mathbb{R}^{N-1}$,
which is the space of {\em non-normalized} texture parameters
(\ref{eq:AppModel}). Retaining all non-zero PCA modes, the
 Euclidean distance between images is exactly retained, also we do not lose any texture
information. Voronoi cell
and distance computations now only have to be performed in $\mathbb{R}^{N-1}$ ($N$ is number of training examples), rather than
$\mathbb{R}^{n}$ ( $n$ is number of pixels). The shape-free texture model is then
built to span the Euclidean space $\mathbb{R}^{N-1}$, which
removes the previous problems as regards representation error
(since this is zero by construction), and also the sub-manifold
curvature errors (the model sub-space is flat by construction).

This is the approach we will take in what follows. To be specific,
in (\ref{eq:SpecSet}), $\mathcal{X}$ is now to be taken as the set
of {\em registered} training images, considered in the reference
frame. $\mathcal{Y}$ is then some large sample set of images,
generated by the shape-free texture model. Note that these sample images need
not be explicitly created in $\mathbb{R}^{n}$, but instead are represented as points in
the space of non-normalized texture parameters $\mathbb{R}^{N-1}$. Specificity computations
 will use either Euclidean ($L_{2}$ distance), or  sum of absolute differences ($L_{1}$, which is more robust to
outliers than the $L_{2}$ distance).

This approach to measuring specificity has fewer theoretical problems. It is also more suited to the task at hand of evaluating registration, whilst the dimensional reduction via PCA makes it a much
more efficient evaluation strategy, suitable for use with large sets of 3D images.

\section{Validation of the Approach}
\label{sec:Validation}

\subsection{Brain Dataset with Ground Truth}


We carried out an initial validation of our NRR evaluation method using spatially perturbed 2D slices taken from a registered set of 3D brain images. We expected that, ideally, the specificity measure would vary monotonically with the degree of perturbation.

Our initial dataset consisted of 274 NRR
T1-weighted 3D MR scans of normal subjects (as in~\cite{Kola}). Not all
images were used in the subsequent evaluation, but the size of the dataset is
important, since the entire dataset was initially registered using a
groupwise method,  hence using all the 274 images gives the  best quality 3D registration.
From these registered 3D images, we extracted mid-brain 2D slices, at an
equivalent level across the set. These slices were cropped,
to produce $141\times 141$ pixel images of the central regions of
the brain.

\begin{figure*}[!t]
\centering
\subfloat[]{\includegraphics[height=0.3\linewidth]{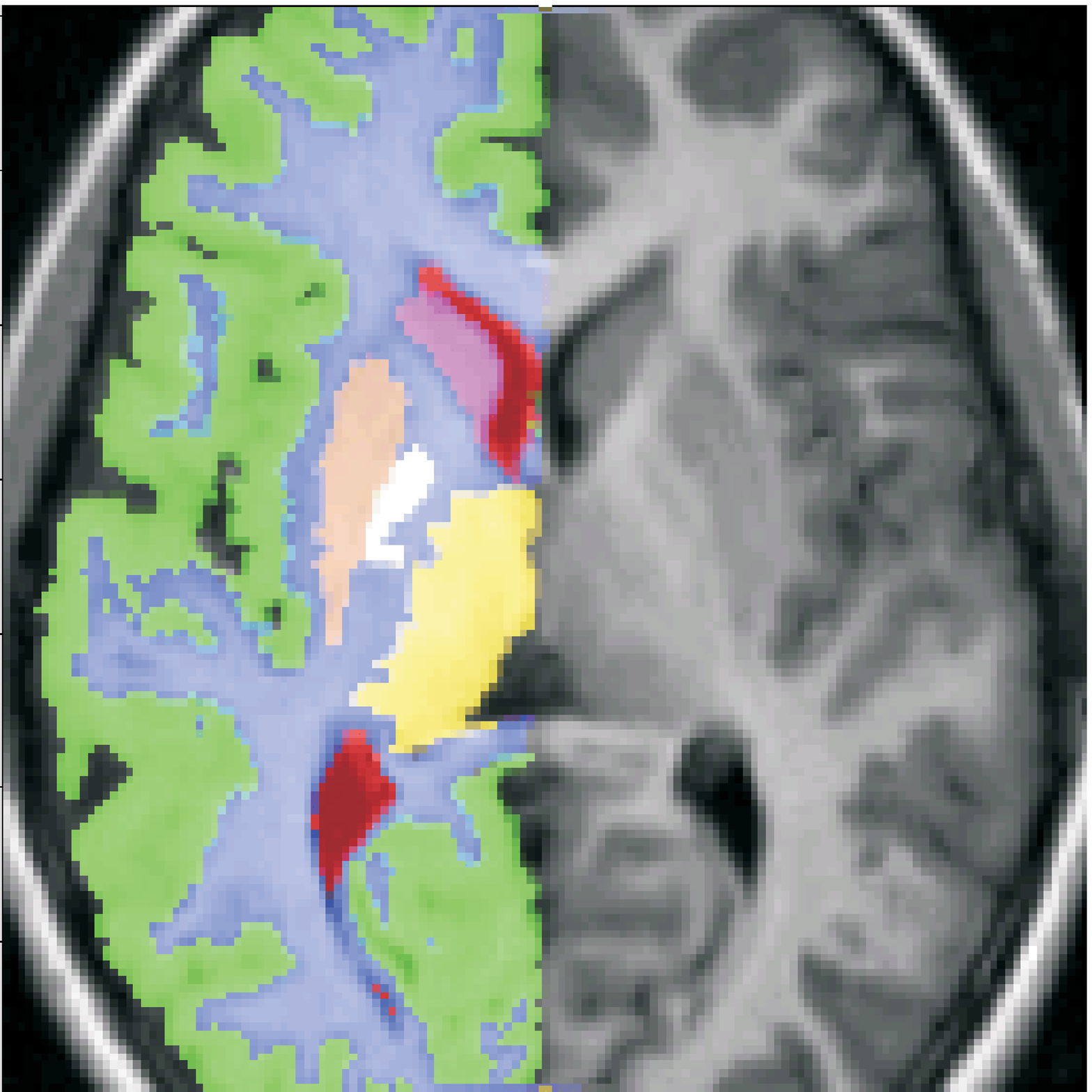}
\label{image_and_labels}}
\hfil
\subfloat[]{\includegraphics[height=0.3\linewidth]{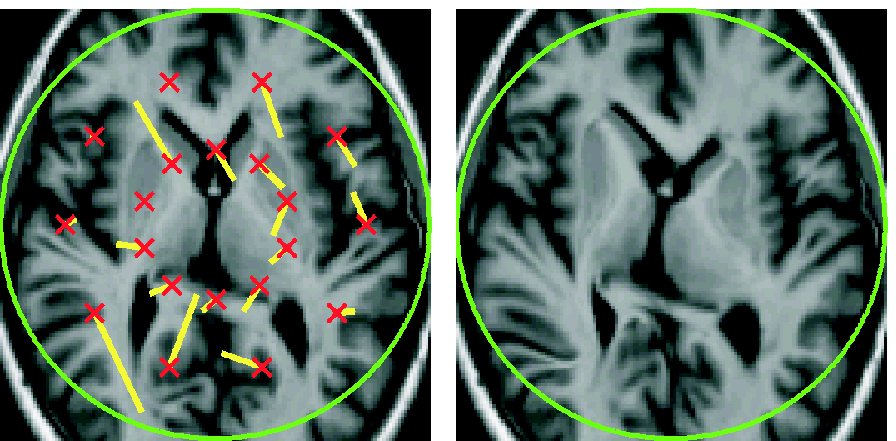}
\label{fig:ExampleBrain}}
\caption{\textbf{Left:} An example registered brain image, overlaid by
its accompanying anatomical labels, for cerebral white matter,
cerebral cortex, lateral ventricle, thalamus, thalamus proper, caudate, and putamen. Labels are also divided into left and
right, one side only shown here. \textbf{Center:} The bounding circle (green), the warp control points (red), and control point displacements (yellow), scaled to show relative not absolute magnitude. \textbf{Right:} The resultant warped image, which has a mean displacement of $d=3.0$ pixels.}
\end{figure*}

The ground truth data for this set consisted of dense (pixel by
pixel) binary tissue labels, the tissue classes being cerebral
white matter, cerebral cortex, lateral ventricle, thalamus,
thalamus proper, caudate, and putamen, also
divided into left and right. Example labels are
shown in Fig.~\ref{image_and_labels}.

\subsection{Perturbing Ground Truth}
We now considered perturbations about
this found registration. Take $I_{i}$ as the $i^{th}$
registered 2D image, sampled at regular pixel positions \mbox{$\{p_{a}:a=1,\ldots 141 \times 141\}$}. A warp of the image plane \mbox{$\psi : \vec{p} \rightarrow \psi(\vec{p})$}
produced a perturbed image function $\tilde{I}_{i}$, where
\mbox{$\tilde{I}_{i}(\psi(\vec{p}_{a})) \doteq
I_{i}(\vec{p}_{a})$}. The values
\mbox{$\{\tilde{I}_{i}(\psi(\vec{p}_{a}))\}$} were
resampled back onto the regular pixel grid $\{\vec{p}_{a}\}$
using bilinear interpolation, to create the perturbed shape-free
texture image \mbox{$\{\tilde{I}_{i}(\vec{p}_{a}):a=1,\ldots 141 \times 141\}$}.

Our non-rigid image warps $\psi_{\mathrm{CPS}}$ were generated using the
biharmonic clamped-plate spline (CPS,~\cite{CPS1,CPS2}).
The CPS has the boundary conditions that the warp
is only non-zero in the interior of the unit circle. When applied
to our square images, we take the boundary circle to be the
inscribed circle (see Fig.~\ref{fig:ExampleBrain}), with a set of
initial knot-points positions arrayed within the circle. This
means that we only deform this central
region of the brain, and not the structure such as the skull. This
makes the task of detecting the perturbations harder, in that
misregistration of the strong tissue-edge features associated
with the skull is much easier to detect than misregistrations of
the more subtle structures within the brain.

To generate a warp with a specified mean pixel displacement $d$, each knot point was moved randomly, then the resultant mean pixel displacement $\hat{d}$ was calculated. Every \textit{pixel} displacement was then scaled by the ratio $d/\hat{d}$, to give the exact required mean pixel displacement. The label images were warped along with the actual images.

\subsection{Validation Results}

\begin{figure}[!t]
\centering
\includegraphics[width=\linewidth]{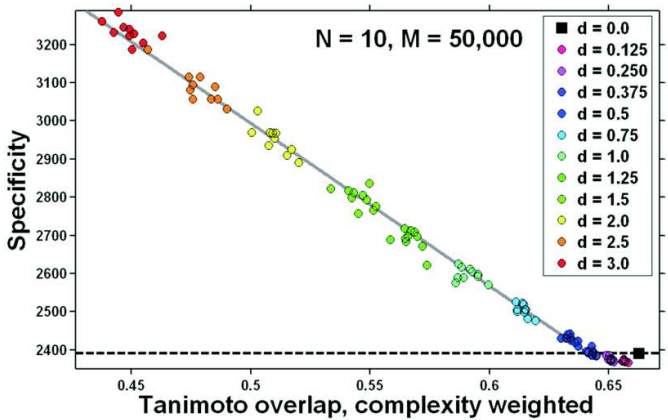}
\caption{Specificity ($M=50,000$) plotted against complexity-weighted Tanimoto overlap, for perturbations of a set of $N=10$ images. The key shows the values of the mean displacement $d$. The dashed horizontal line shows the specificity for zero displacement, whereas the grey line shows the straight-line fit to the data for $d \ge 0.375$.} \label{fig:TanivsSpec.eps}
\end{figure}

We took a fixed subset of $N$ images from the set, and then applied multiple perturbations to this subset. For each instantiation of the perturbation, a value of $d$ was
defined, and a random CPS warp of that size was generated
for each image. We then computed the generalized Tanimoto overlap, and the
specificity for the set of misaligned images. Various weightings were used within the
generalized Tanimoto overlap, and we considered datasets of sizes $10<N\le 50$.
In Fig.~\ref{fig:TanivsSpec.eps}, we show the case of $N=10$
images, subjected to various degrees of perturbation.
We show the complexity-weighted Tanimoto overlap plotted against
the specificity ($S_{\lambda}$ as in (\ref{Seq}), with $M=50,000$
samples, and $\lambda=1$) of the shape-free texture model, for
various values of the mean pixel displacement $d$.

It can be seen that for values of $d \ge 0.375$ pixels, there is a
very good linear relationship between the generalized Tanimoto overlap and the specificity. The
maximum value of $d$ displayed is $d=3.0$, which as can be seen
from Fig.~\ref{fig:ExampleBrain}, is a large warp, but not a
non-diffeomorphic one. Note that for a given value of the mean
displacement $d$, there is some scatter of the points, as we might
have expected given the random nature of the warp-generation
process. The results are analogous when using other weightings, and other subset sizes.

\begin{table}[!t]
\caption{The average size of each label, expressed as a percentage
of the total labelled volume.} \label{tab:labels} \centering
\begin{tabular}{|r|l|}
\hline
         \bf{\textsc{Label}}        & \bf{\textsc{Fractional Volume \%}} \\
\hline
\textsc{cerebral white matter}      & \textbf{ 33.43} \\
\textsc{cerebral cortex}            & \textbf{ 49.26} \\
\textsc{lateral ventricle}          & \textbf{ 0.81} \\
\textsc{inferior lateral ventricle} & \textbf{ 0.05} \\
\textsc{cerebellum white matter}    & \textbf{ 2.22} \\
\textsc{cerebellum cortex}          & \textbf{ 8.33} \\
\textsc{thalamus proper}            & \textbf{ 1.38} \\
\textsc{caudate}                    & \textbf{ 0.50} \\
\textsc{putamen}                    & \textbf{ 0.81} \\
\textsc{pallidum}                   &  \textbf{ 0.28} \\
\textsc{$3^{rd}$ ventricle}         & \textbf{ 0.07} \\
\textsc{$4^{th}$ ventricle}         & \textbf{ 0.13} \\
\textsc{hippocampus}                & \textbf{ 0.57} \\
\textsc{amygdala}                   & \textbf{ 0.29} \\
\textsc{brain stem}                 & \textbf{ 1.69} \\
\textsc{cerebrospinal fluid}        & \textbf{ 0.10} \\
\textsc{accumbens area}             & \textbf{ 0.08} \\
\textsc{left vessel}                & \textbf{ 0.01} \\
\hline
\end{tabular}
\end{table}

The dip in the specificity below the unperturbed value, for sub-pixel displacements $d \le 0.375$, is a result of the image smoothing effect of such small displacements, a consequence of the necessary re-sampling and interpolation. This smoothing tends to move all images very slightly towards the mean image, and hence registers as a \textit{decrease} in the specificity measure. As $d$ increases, the \textit{increases} in the specificity measure from actual misalignment of structures competes with the smoothing effect. It is important to note that this problem at small displacements is not specific to this measure, and that {\em any} measure based on the distribution of the training set, and sensitive to the overall scale of the data set, will suffer from the same problem when evaluated using this perturbation method. Thus, apart from these unavoidable effects, we have shown that the specificity measure is strongly correlated with the ground truth overlap measure, as required.

\section{Assessing and Comparing Registration Algorithms}\label{sec:Experiments}
We now proceed to the reason for defining
these measures, that is, to enable comparison of the performance
of various non-rigid registration algorithms in cases where ground
truth data is not available.

\subsection{Image Data}
The image set to be registered was taken as the first 36 images
from a dataset supplied by the Center for Morphometric Analysis~(CMA) (as in~\cite{Kola}), consisting of T1-weighted 3D MR images of the brain. Images were acquired at different times with
different scanners, and include control subjects,
as well as subjects with Alzheimer's disease, schizophrenia,
attention deficit hyperactivity disorder, and prenatal drug
exposure, with ages ranging from 4 years to 83 years. Manual
annotations are available for 18 separate cortical and sub-cortical structures,
and include both large-scale structures such as white-matter,
cortex, and the ventricles, as well as more subtle structures such
as the amygdala, thalamus, and caudate. A list is
provided in Table~\ref{tab:labels}. This segmentation was
performed at the CMA, using an extensively described
semi-automated protocol~\cite{Filipek,Nishida}. Although the number of images chosen for registration seems
modest, it was sufficient to provide a realistic registration, and
a good separation of the different registration algorithms in terms of label overlap.

\subsection{Registration Algorithms}
Registration algorithms can be divided into two basic classes:
{\em pairwise} (using only a pair of images) and {\em groupwise}.
\subsubsection{\bf Pairwise \& Repeated Pairwise} Given a pairwise algorithm, registration across a \textit{group} of images can then be achieved by successive applications of the algorithm, which we will refer to as \textit{repeated pairwise}, to make clear the distinction
between this and inherently groupwise algorithms (e.g.,~\cite{IPMI_2005_ISBE}).
All images in the group could each be
pairwise-registered to some chosen reference example
(e.g.,~\cite{Rueckert}), but this suffers from the problem
that, in general, the result obtained depends on the choice of
reference. Refinements are possible, but the
important point to note is that the correspondence for a single
training image is defined w.r.t. this reference (which enables
consistency of correspondence to be maintained across the group),
and that the information used in determining the correct
correspondence is limited to that contained in the single training
set image and the single reference image. This approach explicitly does not take
advantage of the full information in the group of images when
defining correspondence~\cite{Cootes_Groupwise_ECCV}. Making
better use of all the available information is the aim of {\em
groupwise} registration algorithms, where correspondence is
determined across the whole set in a principled manner.

A second important issue when constructing a registration
algorithm is the way that the deformation fields are represented
and manipulated. Fluid
registration~\cite{Christensen}, with dense voxel-by-voxel flow over time, allows large-scale diffeomorphic deformations, but increases the computational complexity of the implementation.
However, a fluid-based deformation field can capture subtleties of deformation not
available to other methods. In contrast, a scheme such as piecewise-affine deformation fields
allows a very compact representation of a deformation field, where
the only variables are the positions and displacements of the
nodes of the selected mesh. This compactness of the representation
means that it is suitable for both pairwise and inherently
groupwise registration algorithms. Such piecewise representations
are also suitable for multi-resolution schemes, since the mesh can
easily be sub-divided recursively.

Hence we take as our \textbf{first} registration algorithm a fluid-based
approach, which performs repeated pairwise registration to a single reference image. The output was
a set of dense 3D deformation fields defined at each voxel in each image.

The \textbf{second and third} registration algorithms chosen both used
piecewise-affine deformation fields, and repeated-pairwise
registration to a reference, differing only in their choice of
reference image from the original dataset. These two references
were selected, based on anecdotal evidence, as being close to or
far from the mean of the set of images. The registration itself
was realized as a coarse-to-fine process; a coarse body-centered
cubic tetrahedral mesh was used to define a piecewise affine
deformation field~\cite{PiecewiseAff} over each image in the set.
The deformation fields were optimized in a number of stages,
starting from an initial global alignment achieved with a
three-dimensional affine transformation. In subsequent stages, the
density of the mesh was increased, and locations of individual
mesh points were progressively optimized to match finer-scale
details between each image and the reference. The process used a
sum of absolute image differences as the objective function, and
resulted in a set of optimized landmarks whose locations defined
the final groupwise correspondence across all the images in the
set.

\subsubsection{\bf Groupwise} The \textbf{fourth} method considered was a groupwise registration
approach, in which a combined reference to which each image is
registered is evolved during the process. In this case, the
reference was constructed as the mean image of the registered set.
The reference hence starts blurred due to initial mis-alignment,
but this allows a smooth optimization of deformation fields, the
reference image then sharpening as the set is progressively
brought into finer-scale alignment. The objective function for the
registration is taken to be the sum of absolute differences
between image voxels. In order to eliminate the effects of
resolution, and due to time considerations and the desire to focus
on the evaluation of the broad groupwise approach, we used an
equivalent piecewise affine deformation field representation and
optimization approach to that used in the pairwise cases. A full
description of the algorithm can be found in Cootes et
al.~\cite{CootesPAMI}.

\subsubsection{\bf Affine} Finally, an affine-only registration (that used as initialization
for the piecewise-affine approaches) was included as the \textbf{fifth} method, as a control.

It is important to note that what matters are not the
exact details of the methods of registration, but that we have
applied \textbf{five} different registration methods to the same dataset,
and that (apart from affine-only) each of these can be considered
as a reasonable registration method that might be used in
practise.

\subsection{Ground Truth Label Data and Registration Evaluation}

Ground truth labels encompass the 18 types as detailed in Table~\ref{tab:labels}, and these are not divided into left and right for this experiment. In order to compute the Tanimoto overlap for the registered images
and compare the different registration algorithms, the label
images need to be warped, and the overlaps computed in a {\em
common} spatial reference, as will be detailed in the next
section. More important is the choice of label-weighting~\cite{Crum}
to be used in the computation of
the generalized Tanimoto overlap. In Table~\ref{tab:labels}, we show the
relative volume of each of the 18 labels. It can be seen that
structures such as the cerebral cortex and cerebral white matter
account for over $80\%$ of the volume, with other important
structures having less than $1\%$ of the total volume, with a
range of volumes for the individual structures that encompasses
three orders of magnitude. Hence if we used the simplest choice,
of implicit-volume-weighting
we can see that our evaluation of registration would be
dominated by the registration results achieved on the cerebral
cortex and cerebral white matter.

Hence we instead choose to use inverse-volume weighting, where the
weight for a given label varies as the inverse of
the label volume. This means that in the final measure, when summed over pixels,
each of the labels will carry equal weight, whatever the size of
the individual structure. This choice does not give undue
prominence to any of the labels chosen for inclusion in the
annotation. We also consider the case of simultaneous
inverse-volume weighting {\em and} complexity weighting, to
investigate whether in this case including complexity has any
significant effect on our results. See \cite{Crum} for further details.

\subsection{Building Models and Measuring Specificity}

\begin{table*}
\caption{Generalized Tanimoto overlap  and specificity, measured for the
various registration algorithms shown. The generalized Tanimoto overlap is
presented for two choices of the weighting, inverse-volume
weighted, and inverse-volume weighted combined with a complexity
weighting. Ranking results are presented in terms of rank order,
and quantitative relative rank.} \label{tab:Results} \centering
\setlength{\extrarowheight}{2pt}
\begin{tabular*}{0.735\textwidth}{|b{0.1\textwidth}| *{2}{b{0.1\textwidth}*{2}{|b{0.015\textwidth}|}} *{2}{b{0.1\textwidth}}*{2}{|b{0.015\textwidth}}|}
\hline
 & \multicolumn{6}{c|}{ } & \multicolumn{4}{c|}{ } \\
 & \multicolumn{6}{c|}{\textsc{\large{Tanimoto Overlap}}} & \multicolumn{4}{c|}{\textsc{\large{Specificity}}} \\
\cline{2-11}
 & \multicolumn{3}{p{0.13\textwidth}|}{\textsc{Inverse Volume}} & \multicolumn{3}{p{0.13\textwidth}|}{\textsc{Inverse Volume \& Complexity}} & \multicolumn{1}{c|}{} & \multicolumn{1}{c|}{} & \multicolumn{2}{c|}{} \\ \cline{2-7}
\multicolumn{1}{|p{0.1\textwidth}|}{\textsc{Registration Algorithm}} & \multicolumn{1}{c|}{Score}
 & \multicolumn{2}{c|}{Rank} & \multicolumn{1}{c|}{Score} & \multicolumn{2}{c|}{Rank} & \multicolumn{1}{c|}{Score} & Standard error & \multicolumn{2}{c|}{Rank} \\ \hline
\textsc{Fluid} & \multicolumn{1}{c|}{0.611} & \multicolumn{1}{c|}{1} &\multicolumn{1}{r|}{100.0\%} & \multicolumn{1}{c|}{0.608} & \multicolumn{1}{c|}{1} &\multicolumn{1}{r|}{100.0\%}  & \multicolumn{1}{c|}{0.131} & \multicolumn{1}{c|}{0.0004} & \multicolumn{1}{c|}{1} &\multicolumn{1}{r|}{100.0\%}  \\
\textsc{Groupwise} & \multicolumn{1}{c|}{0.564} & \multicolumn{1}{c|}{2} &\multicolumn{1}{r|}{83.0\%} & \multicolumn{1}{c|}{0.527} & \multicolumn{1}{c|}{2} & \multicolumn{1}{r|}{73.6\%} & \multicolumn{1}{c|}{0.162} & \multicolumn{1}{c|}{0.0005} & \multicolumn{1}{c|}{2} & \multicolumn{1}{r|}{86.8\%} \\
\textsc{Pairwise 1} & \multicolumn{1}{c|}{0.546} & \multicolumn{1}{c|}{4} & \multicolumn{1}{r|}{76.4\%} & \multicolumn{1}{c|}{0.515} & \multicolumn{1}{c|}{3} & \multicolumn{1}{r|}{69.7\%} & \multicolumn{1}{c|}{0.173} & \multicolumn{1}{c|}{0.0006} & \multicolumn{1}{c|}{3} & \multicolumn{1}{r|}{82.2\%} \\
\textsc{Pairwise 2} & \multicolumn{1}{c|}{0.553} & \multicolumn{1}{c|}{3} & \multicolumn{1}{r|}{79.0\%} & \multicolumn{1}{c|}{0.515} & \multicolumn{1}{c|}{3} & \multicolumn{1}{r|}{69.7\%} & \multicolumn{1}{c|}{0.174} & \multicolumn{1}{c|}{0.0005} & \multicolumn{1}{c|}{4} & \multicolumn{1}{r|}{81.8\%}\\
\textsc{Affine} &  \multicolumn{1}{c|}{0.335} & \multicolumn{1}{c|}{5} & \multicolumn{1}{r|}{0.0\%} & \multicolumn{1}{c|}{0.301} & \multicolumn{1}{c|}{5} & \multicolumn{1}{r|}{0.0\%} & \multicolumn{1}{c|}{0.367} & \multicolumn{1}{c|}{0.0010} & \multicolumn{1}{c|}{5} & \multicolumn{1}{r|}{0.0\%}\\ \hline
\end{tabular*}

\end{table*}

For a given set of registered images, both the specificity
(comparing registered training set images), and the generalized Tanimoto
overlap (comparing registered label images) have to be evaluated
in a common spatial reference frame. However, when it comes to
comparing different registrations, it should be noted that in
general, the spatial reference frame of one registration algorithm
need not be the same as the spatial reference frame of another
registration algorithm.

In order for a comparison of registration methods to be
meaningful, we need to map the results of all registration methods
into a single, common spatial reference frame. For the non-fluid
registration methods, correspondence is defined via a set of
mappings from the particular spatial reference frame to the frame
of each original training set image. Whereas in the fluid case,
correspondence is defined indirectly, via a dense mapping from
each individual training image into the fluid reference frame.
Hence when casting all results into a common reference, we use the
fluid spatial reference, since it is straightforward to compose
the mapping from the non-fluid spatial reference to the training
image with the fluid mapping from the training image to the fluid
spatial reference. Hence we obtain, for each registration
algorithm, a registered set of training images (the shape-free
texture images) $\mathcal{X}$, and a registered set of label
images. The generalized Tanimoto overlaps can then be computed.

In order to compute the specificity for each algorithm, we first
need to construct the shape-free texture model for that set of
registered images. Since the spatial frame is common, we can use
the same set of texture sample positions for each registration
algorithm. The shape-free texture model is then constructed using
standard PCA, where we retain {\em all 35} non-normalized texture
parameters/modes of variation in each case.

The sample set $\mathcal{Y}$ for each model is then generated as a
set of $M=500 000$ points in the space of non-normalized model
parameters, distributed according to the distribution of the data
$\mathcal{X}$. The Voronoi cell and distance computations for the
specificity~(\ref{eq:SpecSet}) are then performed in the space of
parameters. We first compute the distances between every sample
point and every training point, to extract the required nearest
training example to each sample point. Note that in this case,
rather than the $L_{2}$ Euclidean distance in parameter space
(equivalent to the Euclidean distance in the space of images), we
instead use the $L_{1}$ sum of absolute differences in parameter
space. Although this $L_{1}$ distance is not rotationally
symmetric, the parameters/axis system used respects the
distribution of the training data, and hence justifies its use.

\subsection{Results}

Table~\ref{tab:Results} gives the results for the specificity and generalized Tanimoto overlap, for all five
methods of registration.

From the rank order results, it can be seen that all measures
agree that the fluid method is best, followed by groupwise,
pairwise, and finally affine; the only disagreement is in terms of
the exact positioning of the two pairwise methods. In terms of the
{\em quantitative relative ranking} (taking the score for the best
method as $100\%$ and the score for the worst as $0\%$, and linear
scaling  between these two extremes), we can see that
inverse-volume weighted generalized Tanimoto overlap is in rough agreement
with the specificity results, with groupwise ranking $>83\%$, and
both pairwise results $\approx 80\%$, with $\approx 4\%$
difference between groupwise and the best of the pairwise. The generalized
Tanimoto overlap evaluated with combined inverse-volume and
complexity weighting gives lowered relative ranking for groupwise
and pairwise, but with still $\approx 4\%$ difference between
groupwise and the best of the pairwise. This is perhaps to be
expected; the addition of the complexity weighting gives more
weight to labels with convoluted boundaries, and the fluid
registration is better able to match such boundaries than the
representation of deformation fields used in the groupwise and
pairwise cases. The affine registration does badly in both cases.
Hence the generalized Tanimoto overlap in the fluid case sees little change
when the complexity weighting is included, whereas the groupwise
and pairwise cases suffer a relative degradation in measured
performance, as is shown  by the relative rankings.

\section{Discussion and Conclusions}
\label{sec:Conclusions}

We have described a model-based approach to assessing the accuracy
of non-rigid registration of groups of images. The most important
thing about this method is that it does not require any
ground truth data, but depends only on the training data itself. A
second important consideration is that the use of {\em registered}
sets of images allows us to map the problem from the
high-dimensional space of images, to the lower-dimensional space
of PCA parameters (whether or not a linear model is then used at
the next stage of actual modelling). This dimensionality-reduction
means that the key figure is the number of examples, not their
dimensionality, which makes the
evaluation of 3D image registration much more practical.


For specificity, extensive validation experiments were conducted,
based on perturbing correspondence obtained through registration.
These show that our method is able to detect increasing
misregistration using just the registered image data.

More importantly, we have shown that what is being measured by our
model-based approach varies monotonically with an overlap measure
based on ground truth, for the case of 2D slices extracted from 3D-registered brain data. Note that it is \textit{not} essential in this case that the misregistration is biologically feasible, since we are not trying to recover the deformation itself, but instead compare the ground-truth based and model-based measures under misregistration.

Finally, we have applied our model-based measure to assessing the
quality of various registration algorithms when applied to 3D MR
brain images. This shows agreement, in terms of both the rank
order and quantitative relative-ranking, between our method and
ground truth based methods of evaluation.

It should be noted that the results presented here for the
relative ranking of the registration algorithms considered are in
general agreement with results presented
elsewhere~\cite{Klein2009,Klein2010}, for extensive ground truth
based evaluation of multiple registration algorithms on brain
images. In particular, in Klein et al.~(2010)~\cite{Klein2010},
the authors considered label-based evaluation of the volumetric
registration of pairs of images, both with and without an
intermediate template. They found that using a customized, average
template, constructed from a group of images similar to the pair
being registered, gave improved performance. This is in agreement
with our ranking of our groupwise algorithm (registration to an evolving averaged reference image) compared to our pairwise algorithms (registration to a {\em
fixed}, non-averaged reference image).

It was noted both by Hellier~(2003)~\cite{Hellier2003}, and by
Klein et al.~(2009)~\cite{Klein2009}, that there is a (modest)
correlation between the number of degrees of freedom of the
deformation and registration accuracy. These observations are in
agreement with our results (given the observation above about the
relative positioning of groupwise compared to pairwise based on
choice of reference or template). That is, the fluid algorithm was
ranked highest (largest number of degrees of freedom), followed by
the closely-spaced groupwise and pairwise methods (the
same mesh-based representation of deformation, with a moderate
number of degrees of freedom), and with affine (very small number
of degrees of freedom) a long way behind.

We here used linear modelling in our
evaluation, but in principle, any generative model-building
approach could be used. This method is, in
principle, very general, and can be applied to the results of any
registration algorithm, as long as the assignment of (anatomical)
correspondence across the entire group of images is both possible
and meaningful.

We have only considered the
registration of images of a single modality, where we obtain a
simple distribution of registered images in image space, clustered
about the mean image. Let us consider briefly the case of
multimodal registration. Suppose we have a group of images
split between two modalities $M_{1}$ and $M_{2}$, where each
modality contains similar anatomical or structural information. We
could then either register the entire set using, for example,
mutual information as an image similarity measure, or we could
register all the $M_{1}$ images, and all the $M_{2}$ images
separately. Either way (provided we make a link between the two
different spatial frames of the two means in the latter case), we
then obtain a distribution of registered images in image space
that will correspond to a cluster of registered images for each
modality. The obvious next step is to model this distribution of
all the images as a gaussian mixture model, allowing us to
compute the specificity of the entire set. The two different registration scenarios presented above could then be
compared. There are obviously other cases of multimodal
registration which can be brought within a similar framework, but
further exploration of this important issue is beyond the scope of
the present paper.

Our model-based method hence represents a significant advance as
regards the important problem of evaluating non-rigid registration
algorithms. It establishes an entirely objective basis for
evaluation, since it is free from the requirement of ground truth
data. It also frees us from one caveat of label-based
evaluation~\cite{Klein2009,Klein2010}, which is that such methods
totally ignore misregistration {\em within} labelled regions.

\section*{Acknowledgement}
The authors would like to thank Prof. D. Kennedy, C.
Haselgrove, and the Center for Morphometric Analysis (CMA) for
providing the MR images used. Also the various groups that
provided the complete labeling data, and K.
Babalola~\cite{Kola}, for his registration results and for processing the images. 


%

\bibliographystyle{IEEEtran}
\bibliography{IEEEabrv,nrr_model_assess}

\begin{thebibliography}{10}
\providecommand{\url}[1]{#1}
\csname url@samestyle\endcsname
\providecommand{\newblock}{\relax}
\providecommand{\bibinfo}[2]{#2}
\providecommand{\BIBentrySTDinterwordspacing}{\spaceskip=0pt\relax}
\providecommand{\BIBentryALTinterwordstretchfactor}{4}
\providecommand{\BIBentryALTinterwordspacing}{\spaceskip=\fontdimen2\font plus
\BIBentryALTinterwordstretchfactor\fontdimen3\font minus
  \fontdimen4\font\relax}
\providecommand{\BIBforeignlanguage}[2]{{%
\expandafter\ifx\csname l@#1\endcsname\relax
\typeout{** WARNING: IEEEtran.bst: No hyphenation pattern has been}%
\typeout{** loaded for the language `#1'. Using the pattern for}%
\typeout{** the default language instead.}%
\else
\language=\csname l@#1\endcsname
\fi
#2}}
\providecommand{\BIBdecl}{\relax}
\BIBdecl

\bibitem{Oliveira_Review}
F.~P. Oliveira and J.~M.~R. Tavares, ``Medical image registration: a review,''
  \emph{Computer Methods in Biomechanics and Biomedical Engineering}, vol.~17,
  no.~2, pp. 73--93, 2014.

\bibitem{Zitova_2003}
B.~Zitov\'{a} and J.~Flusser, ``Image registration methods: A survey,''
  \emph{Image and Vision Computing}, vol.~21, pp. 977 -- 1000, 2003.

\bibitem{Hellier}
P.~{Hellier}, C.~{Barillot}, I.~{Corouge}, B.~{Gibaud}, G.~{Le Goualher}, D.~L.
  {Collins}, A.~{Evans}, G.~{Malandain}, N.~{Ayache}, G.~E. {Christensen}, and
  H.~J. {Johnson}, ``Retrospective evaluation of intersubject brain
  registration,'' \emph{{IEEE} Trans. Med. Imag.}, vol.~22, no.~9, pp.
  1120--1130, 2003.

\bibitem{Validation-NRR}
P.~Rogelj, S.~Kovacic, and J.~C. Gee, ``Validation of a nonrigid registration
  algorithm for multimodal data,'' in \emph{Proceedings of Medical Imaging
  2002, Image Processing, SPIE Proceedings}, vol. 4684, 2002, pp. 299--307.

\bibitem{Schnabel}
J.~A. Schnabel, C.~Tanner, A.~C. Smith, M.~O. Leach, C.~Hayes, A.~Degenhard,
  R.~Hose, D.~L.~G. Hill, and D.~J. Hawkes, ``Validation of non-rigid
  registration using finite element methods,'' in \emph{Information Processing
  in Medical Imaging ({IPMI})}, ser. Lecture Notes in Computer Science,
  M.~Insana and R.~Leahy, Eds., vol. 2082.\hskip 1em plus 0.5em minus
  0.4em\relax Springer, 2001, pp. 344--357.

\bibitem{Klein2009}
A.~Klein, J.~Andersson, B.~A. Ardekani, J.~Ashburner, B.~Avants, M.-C. Chiang,
  G.~E. Christensen, D.~L. Collins, J.~Gee, P.~Hellier, J.~H. Song,
  M.~Jenkinson, C.~Lepage, D.~Rueckert, P.~Thompson, T.~Vercauteren, R.~P.
  Woods, J.~J. Mann, and R.~V. Parsey, ``Evaluation of 14 nonlinear deformation
  algorithms applied to human brain {MRI} registration,'' \emph{NeuroImage},
  vol.~46, no.~3, pp. 786--802, 2009.

\bibitem{Klein2010}
A.~Klein, S.~S. Ghosh, B.~Avants, B.~T.~T. Yeo, B.~Fischl, B.~Ardekani, J.~C.
  Gee, J.~J. Mann, and R.~V. Parsey, ``Evaluation of volume-based and
  surface-based brain image registration methods,'' \emph{NeuroImage}, vol.~51,
  no.~1, pp. 214--220, 2010.

\bibitem{Cootes_2001}
T.~F. Cootes, G.~J. Edwards, and C.~J. Taylor, ``Active appearance models,''
  \emph{{IEEE} Trans. Pattern Anal. Machine Intell.}, vol.~23, pp. 681--685,
  2001.

\bibitem{IPMI_2005_ISBE}
C.~J. Twining, T.~F. Cootes, S.~Marsland, V.~Petrovic, R.~Schestowitz, and
  C.~J. Taylor, ``A unified information-theoretic approach to groupwise
  non-rigid registration and model building.'' in \emph{Proceedings of
  Information Processing in Medical Imaging ({IPMI})}, ser. Lecture Notes in
  Computer Science, G.~Christensen and M.~Sonka, Eds., vol. 3565.\hskip 1em
  plus 0.5em minus 0.4em\relax Springer, 2005, pp. 1--14.

\bibitem{Davies}
R.~H. Davies, C.~J. Twining, T.~F. Cootes, J.~C. Waterton, and C.~J. Taylor,
  ``A minimum description length approach to statistical shape modeling,''
  \emph{{IEEE} Trans. Med. Imag.}, vol.~21, no.~5, pp. 525--537, 2002.

\bibitem{Ashburner2019}
J.~Ashburner, M.~Brudfors, K.~Bronik, and Y.~Balbastre, ``{An algorithm for
  learning shape and appearance models without annotations},'' \emph{Medical
  Image Analysis}, vol.~{55}, pp. {197--215}, {2019}.

\bibitem{goodfellow2014generative}
I.~Goodfellow, J.~Pouget-Abadie, M.~Mirza, B.~Xu, D.~Warde-Farley, S.~Ozair,
  A.~Courville, and Y.~Bengio, ``Generative adversarial nets,'' in
  \emph{Advances in neural information processing systems}, 2014, pp.
  2672--2680.

\bibitem{Jaccard}
P.~Jaccard, ``\'{E}tude comparative de la distribution florale dans une portion
  des {A}lpes et des {J}ura,'' \emph{Bulletin de la Soci\'{e}t\'{e} Vaudoise
  des Sciences Naturelles}, vol.~37, pp. 547--579, 1901.

\bibitem{Tanimoto}
T.~T. Tanimoto, ``An elementary mathematical theory of classification and
  prediction,'' {IBM} Internal Report, 1958.

\bibitem{Tani2}
D.~J. Rogers and T.~T. Tanimoto, ``A computer program for classifying plants,''
  \emph{Science}, vol. 132, no. 3434, pp. 1115--1118, 1960.

\bibitem{Crum}
W.~R. Crum, O.~Camara, and D.~L.~G. Hill, ``Generalized overlap measures for
  evaluation and validation in medical image analysis,'' \emph{{IEEE} Trans.
  Med. Imag.}, vol.~25, no.~11, pp. 1451--1461, 2006.

\bibitem{Cootes_ECCV_1998}
T.~Cootes, G.~Edwards, and C.~Taylor, ``Active appearance models,'' in
  \emph{Proceedings of the European Conference on Computer Vision {(ECCV)}},
  ser. Lecture Notes in Computer Science, H.~Burkhardt and B.~Neumann, Eds.,
  vol. 1407.\hskip 1em plus 0.5em minus 0.4em\relax Springer, 1998, pp.
  484--498.

\bibitem{Edwards}
G.~J. Edwards, T.~F. Cootes, and C.~J. Taylor, ``Face recognition using active
  appearance models,'' in \emph{Proceedings of European Conference on Computer
  Vision ({ECCV})}, ser. Lecture Notes in Computer Science, H.~Burkhardt and
  B.~Neumann, Eds., vol. 1407.\hskip 1em plus 0.5em minus 0.4em\relax Springer,
  1998, pp. 581--595.

\bibitem{de2019quantifying}
M.~de~Groot, N.~Patel, R.~Manavaki, R.~L. Janiczek, M.~Bergstrom,
  A.~{\"O}st{\"o}r, D.~Gerlag, A.~Roberts, M.~J. Graves, Y.~Karkera
  \emph{et~al.}, ``Quantifying disease activity in rheumatoid arthritis with
  the {TSPO PET} ligand {18 F-GE-180} and comparison with {18 F-FDG} and
  {DCE-MRI},'' \emph{{EJNMMI} {R}esearch}, vol.~9, no.~1, pp. 1--11, 2019.

\bibitem{reyneke2018review}
C.~J.~F. Reyneke, M.~L{\"u}thi, V.~Burdin, T.~S. Douglas, T.~Vetter, and T.~E.
  Mutsvangwa, ``Review of {2-D/3-D} reconstruction using statistical shape and
  intensity models and {X-Ray} image synthesis: Toward a unified framework,''
  \emph{IEEE reviews in biomedical engineering}, vol.~12, pp. 269--286, 2018.

\bibitem{cheng2016active}
R.~Cheng, H.~R. Roth, L.~Lu, S.~Wang, B.~Turkbey, W.~Gandler, E.~S. McCreedy,
  H.~K. Agarwal, P.~Choyke, R.~M. Summers \emph{et~al.}, ``Active appearance
  model and deep learning for more accurate prostate segmentation on {MRI},''
  in \emph{Medical Imaging 2016: Image Processing}, vol. 9784.\hskip 1em plus
  0.5em minus 0.4em\relax International Society for Optics and Photonics, 2016,
  p. 97842I.

\bibitem{Frangi}
A.~F. Frangi, D.~Rueckert, J.~A. Schnabel, and W.~J. Niessen, ``Automatic
  construction of multiple-object three-dimensional statistical shape models:
  application to cardiac modelling,'' \emph{{IEEE} Trans. Med. Imag.}, vol.~21,
  pp. 1151--1166, 2002.

\bibitem{US:paper}
C.~J. Twining and C.~J. Taylor, ``Specificity: A graph-based estimator of
  divergence,'' \emph{{IEEE} Trans. Pattern Anal. Machine Intell.}, vol.~33,
  no.~12, pp. 2492--2505, 2011.

\bibitem{US:Book}
R.~Davies, C.~J. Twining, and C.~J. Taylor, \emph{Statistical models of shape:
  optimisation and evaluation}.\hskip 1em plus 0.5em minus 0.4em\relax
  Springer, 2008.

\bibitem{Roy}
R.~Schestowitz, C.~J. Twining, T.~Cootes, V.~Petrovic, C.~J. Taylor, and W.~R.
  Crum, ``Assessing the accuracy of non-rigid registration with and without
  ground truth,'' in \emph{Proceedings of the $3^{rd}$ {IEEE} International
  Symposium on Biomedical Imaging: From Nano to Macro}, 2006, pp. 836--839.

\bibitem{tu2017skeletal}
L.~Tu, M.~Styner, J.~Vicory, S.~Elhabian, R.~Wang, J.~Hong, B.~Paniagua, J.~C.
  Prieto, D.~Yang, R.~Whitaker, and S.~M. Pizer, ``Skeletal shape
  correspondence through entropy,'' \emph{{IEEE} Trans. Med. Imag.}, vol.~37,
  no.~1, pp. 1--11, 2018.

\bibitem{Kola}
K.~O. Babalola, T.~F. Cootes, C.~J. Twining, V.~Petrovic, and C.~Taylor, ``{3D}
  brain segmentation using active appearance models and local regressors,'' in
  \emph{Medical Image Computing and Computer-Assisted Intervention {(MICCAI)}},
  ser. Lecture Notes in Computer Science, D.~Metaxas, L.~Axel, G.~Fichtinger,
  and G.~Sz\'{e}kely, Eds., vol. 5241.\hskip 1em plus 0.5em minus 0.4em\relax
  Springer, 2008, pp. 401--408.

\bibitem{CPS1}
C.~J. Twining and S.~Marsland, ``Constructing an atlas for the diffeomorphism
  group of a compact manifold with boundary, with application to the analysis
  of image registrations,'' \emph{Journal of Computational and Applied
  Mathematics}, vol. 222, no.~2, pp. 411--428, 2008.

\bibitem{CPS2}
S.~Marsland and C.~J. Twining, ``Constructing diffeomorphic representations for
  the groupwise analysis of the nonrigid registrations of medical images,''
  \emph{{IEEE} Trans. Med. Imag.}, vol.~23, no.~8, pp. 1006--1020, 2004.

\bibitem{Filipek}
P.~A. Filipek, C.~Richelme, D.~N. Kennedy, and V.~S. {Caviness Jr}, ``The young
  adult human brain: An {MRI}-based morphometric analysis,'' \emph{Cerebral
  Cortex}, vol.~4, pp. 344--360, 1994.

\bibitem{Nishida}
M.~Nishida, N.~Makris, D.~N. Kennedy, M.~Vangel, B.~Fischl, K.~S.
  Krishnamoorthy, V.~S. Caviness, and P.~E. Grant, ``Detailed semiautomated
  {MRI} based morphometry of the neonatal brain: preliminary results,''
  \emph{NeuroImage}, vol.~32, no.~3, pp. 1041--1049, 2006.

\bibitem{Rueckert}
D.~Rueckert, A.~F. Frangi, and J.~A. Schnabel, ``Automatic construction of
  {3-D} statistical deformation models of the brain using nonrigid
  registration,'' \emph{{IEEE} Trans. Med. Imag.}, vol.~22, no.~8, pp.
  1014--1025, 2003.

\bibitem{Cootes_Groupwise_ECCV}
T.~F. Cootes, S.~Marsland, C.~J. Twining, K.~Smith, and C.~J. Taylor,
  ``Groupwise diffeomorphic non-rigid registration for automatic model
  building,'' in \emph{Proceedings of European Conference on Computer Vision
  (ECCV)}, ser. Lecture Notes in Computer Science, T.~Pajdla and J.~Matas,
  Eds., vol. 3024.\hskip 1em plus 0.5em minus 0.4em\relax Springer, 2004, pp.
  316--327.

\bibitem{Christensen}
G.~E. Christensen, R.~D. Rabbitt, and M.~I. Miller, ``Deformable templates
  using large deformation kinematics,'' \emph{{IEEE} Trans. Image Processing},
  vol.~5, pp. 1435--47, 1996.

\bibitem{PiecewiseAff}
T.~Cootes, C.~Twining, V.~Petrovic, R.~Schestowitz, and C.~Taylor, ``Groupwise
  construction of appearance models using piece-wise affine deformations,'' in
  \emph{Proceedings of the $16^{th}$ {B}ritish {M}achine {V}ision {C}onference
  {(BMVC)}}, vol.~2, 2005, pp. 879--888.

\bibitem{CootesPAMI}
T.~F. Cootes, C.~J. Twining, V.~S. Petrovi\'{c}, K.~O. Babalola, and C.~J.
  Taylor, ``Computing accurate correspondences across groups of images,''
  \emph{{IEEE} Trans. Pattern Anal. Machine Intell.}, vol.~32, no.~11, pp.
  1994--2005, 2009.

\bibitem{Hellier2003}
P.~Hellier, ``{Consistent intensity correction of MR images},'' in
  \emph{Proceedings 2003 International Conference on Image Processing (ICIP)
  {(Cat. No.03CH37429)}}, vol.~1, 2003, pp. 1109--1112.

\end{thebibliography}

\end{document}